\documentclass[a4paper,fleqn]{cas-dc}
\usepackage[authoryear]{natbib}

\usepackage{threeparttable}
\usepackage{flushend}
\usepackage{graphicx}

\usepackage{subcaption}

\def\tsc#1{\csdef{#1}{\textsc{\lowercase{#1}}\xspace}}
\tsc{WGM}
\tsc{QE}

\begin{document}
\begin{sloppypar} 

\let\WriteBookmarks\relax
\def\floatpagepagefraction{1}
\def\textpagefraction{.001}
\let\printorcid\relax

\makeatletter
\renewcommand{\fnum@figure}{Fig. \thefigure.\@gobble}
\makeatother

\shorttitle{}    

\shortauthors{}  

\title [mode = title]{Aerial-ground Cross-modal Localization: Dataset, Ground-truth, and Benchmark}  


\tnotetext[1]{This research was supported by the National Natural Science Foundation Project (No. 42201477, No.42130105)}

\affiliation[1]{organization={Department of Geomatics Engineering},
            addressline={University of Calgary}, 
            city={Calgary},
            postcode={T2N 1N4}, 
            country={Canada}}

\affiliation[2]{organization={School of Electrical and Electronic Engineering},
            addressline={Nanyang Technological University}, 
            postcode={639798}, 
            country={Singapore}}
            
\affiliation[3]{organization={State Key Laboratory of Information Engineering in Surveying, Mapping and Remote Sensing},
            addressline={Wuhan University}, 
            city={Wuhan},
            postcode={430079}, 
            country={China}}

\affiliation[4]{organization={School of Mechanical Engineering},
            addressline={Chongqing Technology and Business University}, 
            city={Chongqing},
            postcode={400067}, 
            country={China}}

\author[1]{Yandi Yang}

\ead{yandi.yang@ucalgary.ca}

\credit{Conceptualization of this study, Methodology, Software, Validation, Writing – original draft}

\author[2]{Jianping Li}
\cormark[1]
\ead{jianping.li@ntu.edu.sg}
\credit{Conceptualization of this study, Methodology, Writing review}

\author[3]{Youqi Liao}
\ead{martin_liao@whu.edu.cn}
\credit{Methodology, Writing review}

\author[4]{Yuhao Li}
\ead{yhaoli@whu.edu.cn}
\credit{Methodology, Writing review}

\author[3]{Yizhe Zhang}
\ead{yizhezhang0418@whu.edu.cn}
\credit{Data collection, Writing review}

\author[3]{Zhen Dong}
\ead{dongzhenwhu@whu.edu.cn}
\credit{Methodology, Writing review}

\author[3]{Bisheng Yang}
\ead{bshyang@whu.edu.cn}
\credit{Conceptualization of this study, Writing review}

\author[1]{Naser El-Sheimy}
\ead{elsheimy@ucalgary.ca}
\credit{Conceptualization of this study, Methodology, Writing review, Project administration, Funding acquisition}
\cortext[1]{Corresponding author}



\begin{abstract}
Accurate visual localization in dense urban environments poses a fundamental task in photogrammetry, geospatial information science, and robotics. While imagery is a low-cost and widely accessible sensing modality, its effectiveness on visual odometry is often limited by textureless surfaces, severe viewpoint changes, and long-term drift. The growing public availability of airborne laser scanning (ALS) data opens new avenues for scalable and precise visual localization by leveraging ALS as a prior map. However, the potential of ALS-based localization remains underexplored due to three key limitations: (1) the lack of platform-diverse datasets, (2) the absence of reliable ground-truth generation methods applicable to large-scale urban environments, and (3) limited validation of existing Image-to-Point Cloud (I2P) algorithms under aerial-ground cross-platform settings.
To overcome these challenges, we introduce a new large-scale dataset that integrates ground-level imagery from mobile mapping systems with ALS point clouds collected in Wuhan, Hong Kong, and San Francisco. Accurate 6-DoF ground-truth poses for the images are obtained indirectly, by first aligning mobile LiDAR scans (MLS) with ALS data through ground segmentation, façade reconstruction, and multi-sensor pose graph optimization. Given the rigid mounting between LiDAR and camera, the optimized MLS trajectory is then transferred to the image stream, enabling reliable image-ALS localization. This strategy circumvents the difficulty of direct cross-modal registration and provides a robust basis for benchmarking. Finally, we evaluate state-of-the-art global and fine I2P localization methods under cross-view and cross-modal conditions. Our dataset and benchmarks aim to facilitate research in air–ground localization, multi-source data fusion, and urban-scale photogrammetric applications.  The project page is available at: \url{https://yandiyang.github.io/aerial_ground_cross_modal/}.
\end{abstract}








\maketitle

\section{Introduction}




Accurate localization in urban environments is a core task in photogrammetry, geospatial information science, and robotics, underpinning a wide range of applications such as urban mobile mapping \citep{schwarz2004mobile}, autonomous navigation \citep{el2020inertial}, and robotic delivery \citep{li2024hcto,li2025graph}. In densely built-up areas, Global Navigation Satellite System (GNSS) signals are often severely degraded or entirely unavailable due to signal occlusion and multipath effects caused by tall structures and narrow streets \citep{nassar2006combined}. Although Inertial Navigation Systems (INSs) can maintain short-term localization continuity, they suffer from cumulative drift in the absence of reliable external corrections \citep{el2020inertial}.
Among various sensing modalities, consumer-grade cameras have become the most ubiquitous due to their low cost, compact form factor, and widespread integration across mobile platforms, including smartphones, vehicles, and autonomous systems. As a passive, power-efficient, and semantically rich sensing modality, imagery has strong potential for scalable localization in complex urban scenes with prebuilt reference maps, such as point clouds \citep{liSaliencyI2PLocSaliencyguidedImage2025,zou2025reliable,liao2024mobile}, ground view images \citep{warburg2020mapillary}, airborne images \citep{jende2018fully}, and OpenStreetMap (OSM)\citep{liao2024osmloc}.

\begin{figure}
\centering
\includegraphics[width=\linewidth, keepaspectratio]{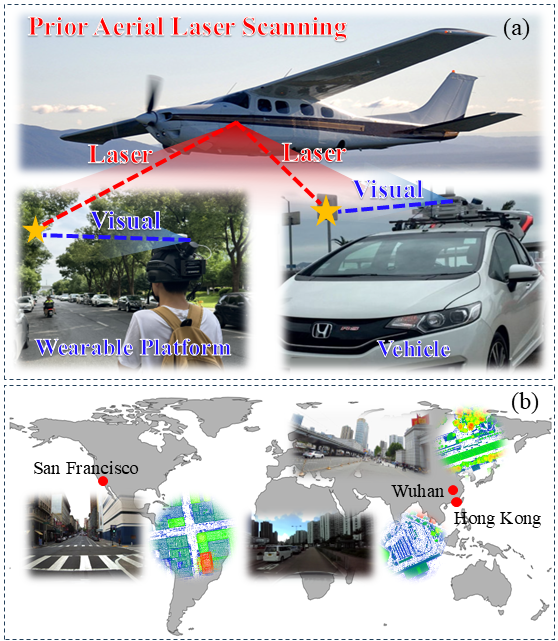}
\caption{Global distribution of our dataset on the map.}
\label{fig_map}
\end{figure}

Visual localization using pre-built maps in highly urbanized environments remains a challenging task. Although numerous visual place recognition algorithms have been proposed, most existing ground-view image references, such as those from Mapillary, are collected non-uniformly by tourists or ground vehicles, resulting in sparse and uneven spatial coverage \citep{warburg2020mapillary}. Relocalization based on 2D reference maps, including satellite imagery and OpenStreetMap (OSM), is limited to 3-DoF pose estimation with restricted accuracy, despite their broad accessibility and ease of use \citep{liao2024osmloc,sarlin2023orienternet,sarlin2023snap}. In contrast, 6-DoF localization can be achieved using 3D pre-built maps such as point clouds and meshes, offering improved geometric fidelity. Furthermore, the growing public availability \citep{R89,R90} of Airborne Laser Scanning (ALS) point clouds from governmental agencies worldwide presents new opportunities for scalable and accurate 6-DoF visual localization using ALS as a prior map.

Despite its potential, ALS-based visual localization in urban environments remains underexplored due to three major challenges. First, most publicly available datasets lack platform diversity, particularly the integration of both ground and aerial sensing platforms. To date, there is no dataset specifically designed to support image-based localization using ALS data in urban contexts.
Second, ground-truth trajectories are often obtained through loop closure techniques or depend on external reference systems such as high-end IMUs \citep{r38}. However, these approaches become unreliable or impractical for ground-based platforms operating in large-scale urban environments \citep{el2020inertial}.
Third, existing Image-to-Point Cloud (I2P) methods are predominantly developed for ground-platform scenarios \citep{liSaliencyI2PLocSaliencyguidedImage2025}, limiting their generalizability across aerial and ground platforms. 

The above challenges underscore the need for a comprehensive benchmark that facilitates rigorous evaluation of cross-view and cross-modal localization methods. To address these issues, we propose an air-ground cross-modal localization dataset. The main contributions of this article are as follows.

\begin{enumerate}[1)]
\setlength{\leftmargin}{0em}


\item We present a new large-scale dataset that enables aerial-ground cross-modal localization by combining ground-level imagery from mobile mapping systems with ALS point clouds. The data span three representative urban areas—Wuhan, Hong Kong, and San Francisco—and will be made publicly accessible to the research community.

\item We propose an indirect yet scalable approach for generating accurate 6-DoF ground-truth image poses. This is achieved by registering mobile LiDAR submaps to ALS data using ground segmentation and façade reconstruction, followed by multi-sensor pose graph optimization.

\item We establish a unified benchmarking suite for both global and fine-grained I2P localization, and evaluate state-of-the-art methods under challenging cross-view and cross-modality conditions. Future research trends are summarized according to the evaluation results.
\end{enumerate}

The remainder of this paper is organized as follows. Section~\ref{sec_2} reviews existing algorithms and datasets related to cross-view and cross-modal localization, along with methods for generating ground-truth trajectories. Section~\ref{sec_3} details the composition and characteristics of the proposed dataset. Section~\ref{sec_4} describes the methodology for ground-truth trajectory generation, while Section~\ref{sec_5} presents its qualitative and quantitative evaluation. Section~\ref{sec_6} benchmarks state-of-the-art Image-to-Point Cloud (I2P) localization algorithms under aerial-ground cross-modal settings. Section~\ref{sec_7} discusses the remaining challenges in air–ground cross-modal localization. Finally, conclusions are drawn in Section~\ref{sec_8}.

\section{Related work}
\label{sec_2}

\subsection{Cross-platform localization and dataset}
2D maps, such as satellite images \citep{ye2024cross} and OpenStreetMap \citep{wu2024maplocnet}, have been widely used for visual localization due to their extensive coverage and accessibility. However, they are restricted to estimating 3-DoF poses with limited accuracy. Various forms of 3D maps can also be used to enhance localization accuracy, with point clouds \citep{zhang2023cross} being a widely adopted choice. However, acquiring prior point cloud maps is often costly and time-consuming, which limits their broad applicability. A similar challenge arises with methods that rely on 3D line maps \citep{r74}, 3D meshes \citep{r76}, CAD models \citep{R77}, and HD maps \citep{r80}, as data acquisition also demands significant resources. Although DEMs (Digital Elevation Models) are mostly publicly available by government agencies, their coarse resolution limits their localization accuracy \citep{r82}.

SLAM datasets incorporating both aerial and ground platforms \citep{R83,R47,R49,R48,R50}, as well as multiple ground-based platforms \citep{r41,r85}, have been published. Nevertheless, these datasets typically cover only limited areas. CS-CAMPUS3D \citep{R86} is a dataset for point cloud place recognition task with self-collected ground LiDAR scans and aerial LiDAR scans from the state government. However, this task can only yield coarse localization results. D-GLSNet \citep{R87} provides a dataset of satellite images aligned with point clouds from KITTI \citep{R88} for 2D-3D matching. Nonetheless, the matching results are limited to 3-DoF poses, providing only horizontal positions and a heading angle.

\subsection{Cross-modal localization and dataset}
Many Image-to-Point Cloud (I2P) algorithms have been developed to solve the cross-modal localization problem. Deep learning techniques have been used to build correspondences between image key points and point clouds. P2-Net \citep{wang2021p2} matches pixel and point directly by jointly learned feature description. CoFiI2P \citep{kang2024cofii2p} proposed a coarse-to-fine pipeline for progressive alignment. DeepI2P \citep{li2021deepi2p} converts the registration problem to a classification task to avoid building correspondences between two modalities. Extrinsic calibration between LiDARs and cameras is a task for accurate registration results. Some solutions are constrained by artificial targets like Apriltags \citep{R19}, and checkerboards \citep{ R17}. For targetless calibration, geometric features like lines \citep{R30}, edges \citep{ R29}, planes \citep{R23} and vanishing points \citep{R24} have been exploited. Extrinsics can also be solved after ego-motion of each sensor has been estimated \citep{R28}. Semantic features are considered for both point clouds and images\citep{R35}. 

Visual SLAM in prior point cloud maps is another task in cross-modal localization. Feature-based methods establish constraints between 2D image and 3D maps features. Line features are extracted from both images and point clouds and subsequently matched in a VIO system \citep{zheng2024tightly}. HyperMap\citep{chang2021hypermap} leverages 3D sparse convolution to extract and compress features for the point cloud map. Registration-based methods typically align image point clouds with prior LiDAR maps \citep{zuo2019visual}. Depth images from point cloud maps and stereo cameras can be aligned for localization \citep{kim2018stereo}. Semantic information such as lane lines and road markings can also be applied for visual localization \citep{qin2021light}. 

In general, current cross-modal localization solutions predominantly operate on single-platform datasets, such as those captured from cars \citep{R36, R37}, backpacks \citep{R46}, handheld devices \citep{R44,R45}, helmets \citep{r38}, UAVs \citep{R39,R40,R43}, and UGVs \citep{R42,r41}. The applicability to datasets from aerial and ground platforms—such as GRACO \citep{R47},M3ED \citep{R49}, SubT-MRS \citep{R48} and CoPeD \citep{R50}-remains constrained.

\subsection{Ground truth trajectory Generation}
Ground truth trajectories serve as a benchmark for SLAM evaluation \citep{r38} and deep learning training \citep{kang2024cofii2p}. Methods that directly refine and optimize maps, such as photogrammetry \citep{R59} and LiDAR scan matching \citep{R58}, derive ground truth trajectories by enhancing map consistency. However, they rely on revisited areas, the availability of which is environment-dependent, making the ground truth generation variable across different scenarios. Beyond leveraging map features, ground truth trajectories can also be acquired using external reference systems, including RTK \citep{R40}, RTK/INS\citep{R54}, motion capture system \citep{R55}, total station\citep{R42,R39}, control point\citep{R56}, TLS point clouds \citep{R48,R57} and fiducial marker\citep{R50}. Nevertheless, these methods are laborious for large-scale areas and restricted by visibility conditions. In addition, ground truth trajectories may become less reliable due to GNSS multipath effects \citep{R54,R60}, failures in LiDAR scan matching \citep{R45}, and the diminishing effectiveness of ground surveys for large-scale trajectory refinement\citep{R61} in real-world scenarios. Our method indirectly generates ground-truth poses by registering mobile laser scanning (MLS) submaps to pre-georeferenced airborne laser scanning (ALS) point clouds and optimizing a pose graph. Leveraging the global coverage of ALS, it avoids reliance on revisited areas and significantly reduces labor.

\section{Dataset description}
\label{sec_3}

\begin{figure*}
\centering
\includegraphics[width=0.9\textwidth, keepaspectratio]{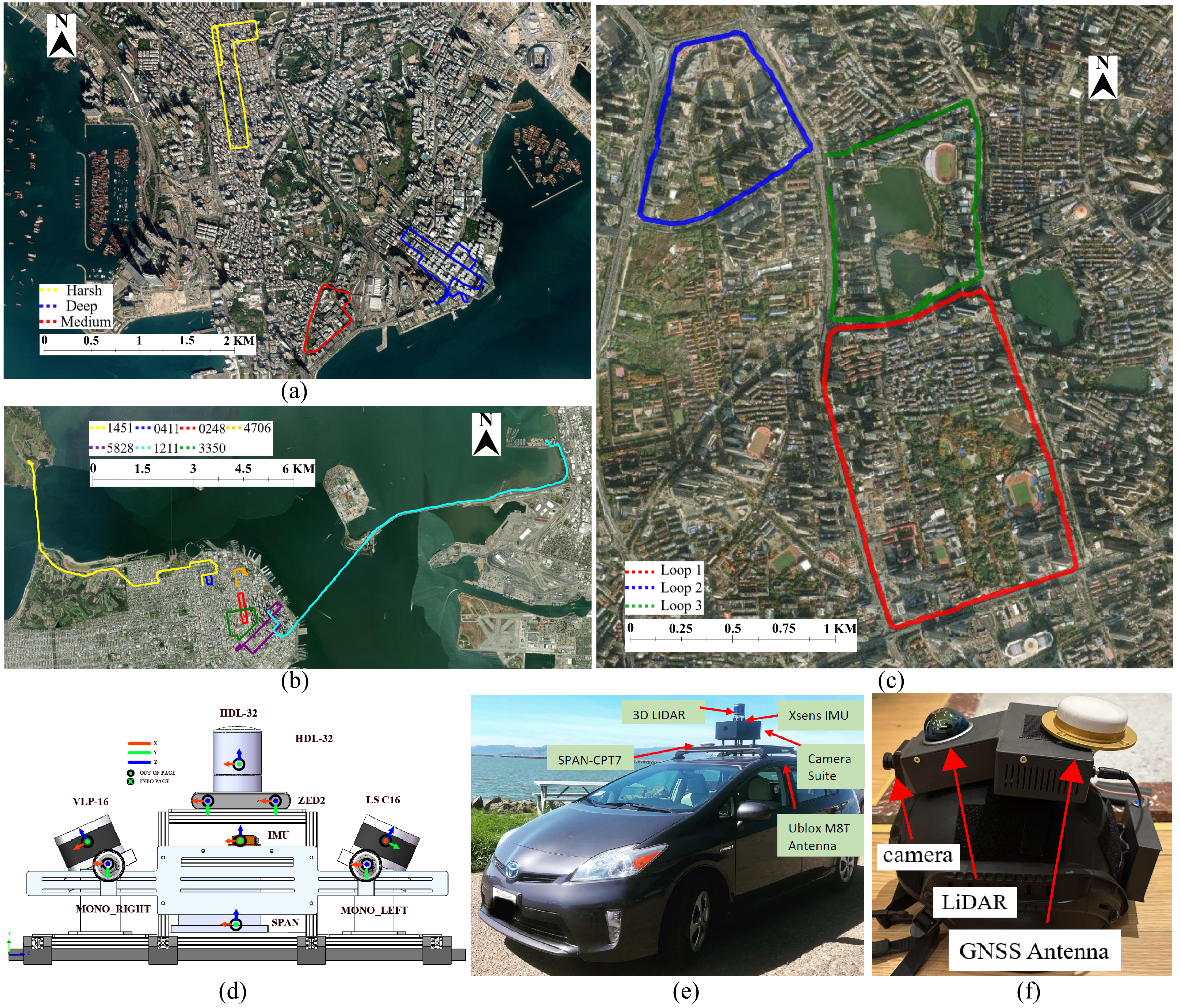}
\caption {Dataset coverage and collection. (a), (b) and (c) illustrate trajectories of the Hong Kong, California and Wuhan datasets, while (d), (e) and (f) are the corresponding data acquisition platforms, with (d) and (e) provided by authors of UrbanNav \citep{R54} and UrbanLoco \citep{R60}, respectively.}
\label{fig_urbannav_traj}
\end{figure*}

\begin{table}[!t]
\centering
\caption{Number of image–ALS point cloud pairs for each sequence across the three datasets.}
\label{tab_all_counts}
{\small
\begin{tabular*}{\linewidth}{@{\extracolsep{\fill}}lcc}
\toprule
Dataset     & Sequence     & Pairs \\
\midrule
\multirow{7}{*}{California} 
            & 3350         & 4{,}960 \\
            & 0248         & 3{,}923 \\
            & 0411         & 1{,}384 \\
            & 4706         & 1{,}611 \\
            & 5828         & 6{,}943 \\
            & 1211         & 7{,}780 \\
            & 1451         & 7{,}082 \\
\midrule
\multirow{3}{*}{Hong Kong}  
            & Deep         & 6{,}579 \\
            & Harsh        & 6{,}870 \\
            & Medium       & 3{,}137 \\
\midrule
\multirow{3}{*}{Wuhan}      
            & Loop 1       & 8{,}526 \\
            & Loop 2       & 5{,}206 \\
            & Loop 3       & 5{,}663 \\
\bottomrule
\end{tabular*}
}
\end{table}

Our dataset integrates ground-level imagery and aerial laser scanning (ALS) data collected in three major urban cities—Hong Kong, San Francisco, and Wuhan—using various mobile mapping systems. This section outlines the geographic coverage, sensing modalities, and platform configurations for each region.

\subsection{Study area}
As shown in Figure ~\ref{fig_map}, the Hong Kong, San Francisco, and Wuhan datasets are selected for their unique advantages. All datasets cover both commercial cores and residential neighborhoods. For instance, the San Francisco dataset includes data collected from Market Street, a major transit artery, and Russian Hill, which reflects typical residential areas. Similarly, the Hong Kong dataset covers Tsim Sha Tsui and Nathan Road, two of Hong Kong's busiest commercial districts, as well as Whampoa, a large-scale residential zone. The Wuhan dataset focuses on the Hankou area, which contains dense urban blocks and bustling commercial streets. In addition, a variety of terrains are included, such as flat areas (e.g., Tsim Sha Tsui), hilly regions (e.g., Russian Hill), and elevated infrastructure (e.g., the Golden Gate Bridge). Lastly, these three cities are representative global megacities for their high urban densities, making the datasets valuable for cross-view and cross-modal localization tasks. The trajectories and data acquisition platforms are illustrated in Figure ~\ref{fig_urbannav_traj}.

\subsection{Ground platform}

The ground datasets leveraged in this paper originate from UrbanNav \citep{R54} and UrbanLoco \citep{R60}, both of which offer multimodal data (i.g. LiDAR, camera, IMU, GNSS and INS/GNSS) operating in highly-urbanized environments.

\textbf{UrbanNav \citep{R54}}: The Hong Kong subset of the UrbanNav dataset has been selected. To construct our dataset, only the horizontal LiDAR (HDL-32E Velodyne) and the ZED2 camera are utilized. The reference trajectory is derived from SPAN-CPT. The tunnel dataset is excluded due to LiDAR degeneration and the lack of aerial data coverage. The combined trajectory length of the remaining three datasets—Medium, Deep, and Harsh—exceeds 13 kilometers.

\textbf{UrbanLoco \citep{R60}}: The San Francisco subset of the UrbanLoco dataset has been selected, featuring a RoboSense 32-line LiDAR and six 360-degree view cameras. The reference trajectory is obtained from SPAN-CPT. Collectively, the seven datasets cover a total trajectory length of over 35 kilometers.

\textbf{WHU-Helmet \citep{r38}}: We collected three trajectory loops in Wuhan using WHU-Helmet, a helmet-based mobile mapping system equipped with a Mid360 LiDAR, a camera, and a GNSS antenna. The data were mainly acquired along sidewalks in dense urban environments, with a total trajectory length exceeding 11 km.

\subsection{Aerial platform}
The aerial data incorporated into our dataset consists of publicly available ALS point clouds obtained from government agencies in Hong Kong \citep{R89} and San Francisco \citep{R90}. These 3D point clouds are acquired using airborne LiDAR sensors, making them particularly valuable for 6-DoF localization tasks compared to 3-DoF localization methods that rely on 2D maps (e.g., satellite imagery, OpenStreetMap).

\textbf{Hong Kong ALS data \citep{R89}}: Acquired in 2020, the Hong Kong ALS point clouds \citep{R89} have an average point spacing of 0.25 meters. Accuracy assessments based on checkpoints indicate a vertical precision of 0.1 meters on flat open terrain, while the horizontal accuracy is measured at 0.3 meters.

\textbf{San Francisco ALS data \citep{R90}}: San Francisco B23 LiDAR \citep{R90} was collected in 2023, with an average point spacing of 0.15 meters. Based on 30 checkpoints in non-vegetated areas, the vertical accuracy is 0.196 meters. Taking into account factors such as flying altitude, IMU error, and GNSS positional error, the horizontal accuracy is calculated to be 0.12 meters.

\textbf{Wuhan ALS data}: Collected in 2024, the ALS point clouds of Wuhan cover densely urbanized areas of Hankou, with an average point spacing of 0.15 m and an average horizontal accuracy of 20 cm.

\begin{figure}[]
\centering
\includegraphics[width=\linewidth, keepaspectratio]{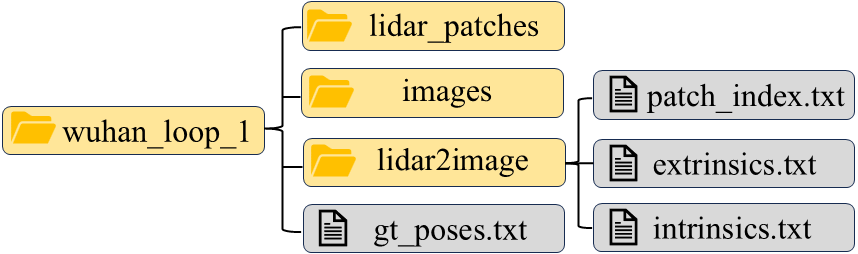}
\caption{File structure of the dataset (taking Wuhan Loop 1 as an example).}
\label{fig_file}
\end{figure}

\subsection{Data overview}
The file structure of the dataset is illustrated in Figure ~\ref{fig_file}. Each sequence contains four components: a folder of aerial LiDAR (ALS) patches, a folder of ground images, a text file containing ground-truth poses, and a folder for projecting ALS point clouds onto the ground images. Table~\ref{tab_all_counts} demonstrates the distribution of image–ALS point cloud pairs across sequences.

\textbf{ALS patches:} All aerial LiDAR point clouds are downsampled to a resolution of 0.5 meters. To improve processing efficiency, the downsampled point clouds are further divided into patches of 100~m\textsuperscript{2} each.

\textbf{Images:} To reduce computational load, images from Wuhan and California are downsampled by 50\% and 25\%, respectively. Images are sampled along the trajectory at intervals of 0.5 meters, and each image corresponds to a specific LiDAR patch.

\textbf{ALS to image projection:} To enable the projection of ALS patches onto ground images, the dataset provides the extrinsic calibration between the ground LiDAR and camera, as well as the camera intrinsics. In addition, the file \texttt{patch\_index.txt} records the pairing between each image (\texttt{*.jpg}), its corresponding ALS patch (\texttt{*.las}), and the center coordinate of that patch.

\textbf{Ground-truth trajectory:} Each line in the pose file (\texttt{.txt} format) contains 16 values representing a 4×4 transformation matrix. These poses describe the ground LiDAR positions in the mapping frame (ENU frame). To project a 3D ALS point onto the image plane, the following equation is used:

\begin{equation}
\mathbf{p}_{\text{img}} = \mathbf{P} \mathbf{T}_{\text{ext}} \mathbf{T}^{-1} \mathbf{p}_{\text{als}},
\end{equation}
where \( \mathbf{p}_{\text{als}} \) is a homogeneous ALS point, \( \mathbf{T} \) is the ground LiDAR pose, \( \mathbf{T}_{\text{ext}} \) is the extrinsic matrix from ground LiDAR to camera, and \( \mathbf{P} \) is the camera intrinsic matrix.

\section{Ground truth generation}
\label{sec_4}

\begin{figure}[]
\centering
\includegraphics[width=\linewidth, keepaspectratio]{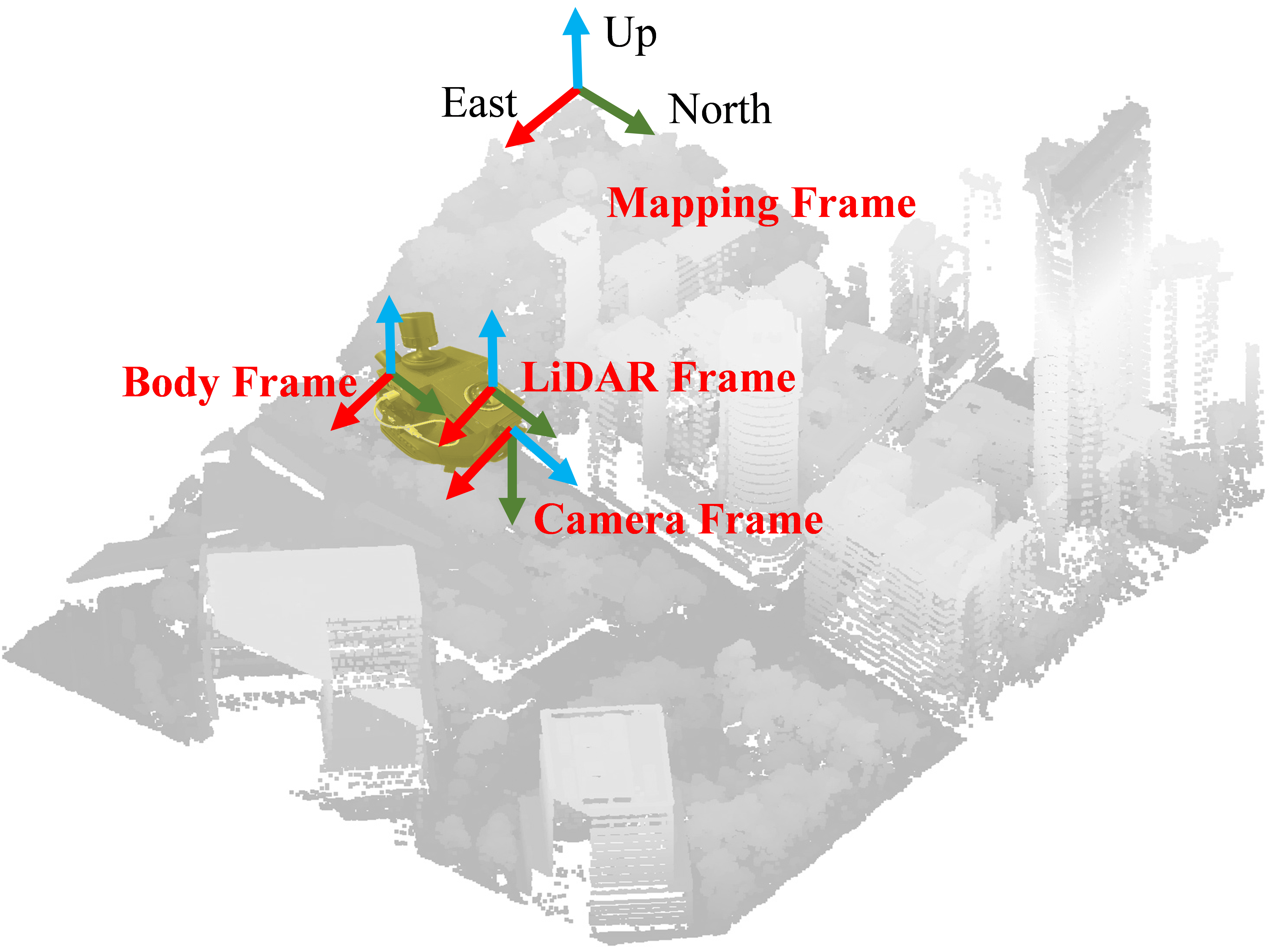}
\caption{Definition of coordinate frames.}
\label{fig_frames}
\end{figure}

To obtain accurate 6-DoF ground-truth poses for images, we adopt an indirect strategy: we first align Mobile Laser Scanning (MLS) data from the vehicle platform with ALS point clouds. Since the MLS and camera are rigidly mounted on the same platform, the resulting refined MLS trajectory serves as a proxy to provide accurate image poses by applying calibration parameters.
To this end, system states are initialized by aggregating all LiDAR frames. Feature correspondences are then extracted—ground segmentation is applied to MLS data, while roof structures are extracted from ALS point clouds to complete building facades. These multi-source constraints are integrated into a factor graph, through which the optimized MLS trajectory, and hence the image ground-truth poses, are obtained. Figure~\ref{fig_frames} illustrates the four coordinate frames involved in the ground-truth optimization pipeline, including the mapping frame, LiDAR frame, camera frame, and IMU (body) frame.


\subsection{Formulation}
Let $\mathbf{X}_k$ denote the state of the $k$-th point cloud frame. The full state vector is defined as follows:

\begin{align}
\mathcal{X} &= \left[ \mathbf{X}_0, \mathbf{X}_1, \dots, \mathbf{X}_n \right], \\
\mathbf{X}_k &= \left[ \mathbf{t}_k, \mathbf{v}_k, \mathbf{\omega}_k, \mathbf{b}_k^a, \mathbf{b}_k^g \right], \quad k \in [0, n],
\end{align}
where $\mathbf{t}_k$, $\mathbf{v}_k$, and $\mathbf{\omega }_k$ represent the position, velocity, and Lie algebra of rotation matrix $\mathbf{R}_k \in \mathrm{SO}(3)$, respectively. $\mathbf{b}_k^a$ and $\mathbf{b}_k^g$ denote the accelerometer and gyroscope biases. $\mathbf{n}$ is the total number of states.

\subsection{Correspondences extraction between MLS and ALS}

\begin{figure*}
\centering
\includegraphics[width=0.9\textwidth, keepaspectratio]{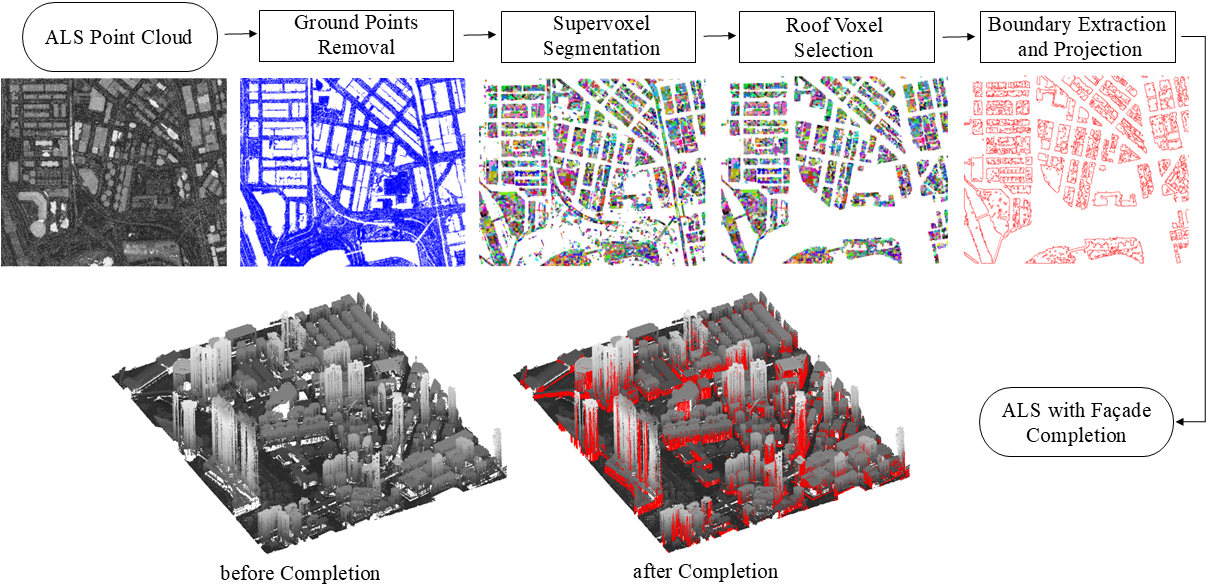}
\caption {The flowchart of ALS feature extraction. Gray points represent the original ALS point clouds (rendered in grayscale based on elevation values) while red points indicate the projected façade points.}
\label{fig_facade}
\end{figure*}

To select salient features for MLS and ALS matching, the MLS point clouds are classified into ground and non-ground points. For each MLS scan, points having the highest planarity \citep{R92} and the lowest verticality \citep{R91} are identified as initial ground plane seeds, which are then used to fit the ground plane using the RANSAC \citep{R93} algorithm. 

Ground points in ALS, which help to correct vertical position errors as well as roll and pitch angles in the MLS trajectory, have already been classified. To further eliminate errors in the horizontal position and heading angle of the MLS trajectory, we leverage ALS building point clouds as reliable features. To extract building point clouds, non-ground points are first segmented into supervoxels \citep{R94}, which are then merged with neighboring supervoxels based on voxel normals and centroids. Supervoxels with planarity \citep{R92} greater than 0.5 and verticality \citep{R91} less than 0.3 are classified as building roof points. To address gaps in the ALS point clouds on building facades, roof boundaries are extracted using a convex hull and projected onto the ground to complete the facade points. Figure~\ref{fig_facade} demonstrates the workflow of ALS feature extraction, where the point clouds are enhanced with façade points.

\subsection{Pose graph optimization}

\begin{figure*}
\centering
\includegraphics[width=0.9\textwidth, keepaspectratio]{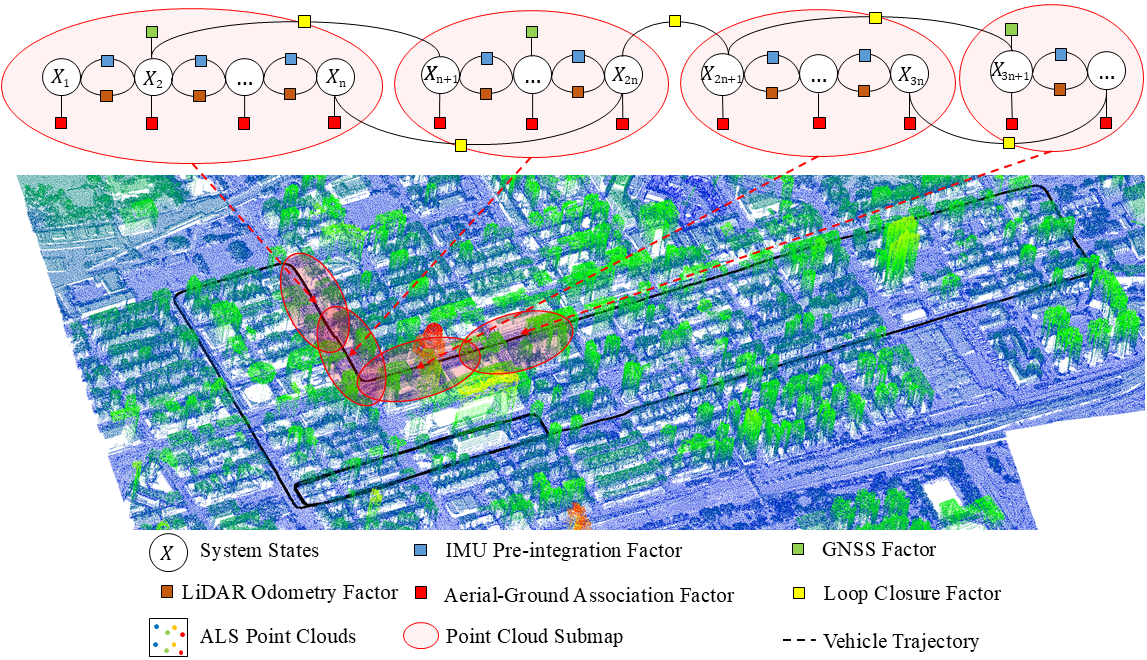}
\caption{Pose graph formulation for ground-truth generation.}
\label{fig_pgo}
\end{figure*}

To obtain high-precision ground-truth trajectories, we formulate a tightly coupled multi-sensor pose graph optimization framework. Each node in the graph represents the complete state of a LiDAR frame, including the platform's pose, velocity, and inertial measurement unit (IMU) biases. The full state vector is defined as:
\begin{align}
\mathcal{X} = \{ \mathbf{X}_0, \mathbf{X}_1, \ldots, \mathbf{X}_N \}, \quad 
\mathbf{X}_i = (\mathbf{R}_i, \mathbf{t}_i, \mathbf{v}_i, \mathbf{b}_i^a, \mathbf{b}_i^g).
\end{align}

The cost function is defined as the weighted sum of residuals from five types of factors:
\begin{align}
\min_{\mathcal{X}} \quad 
&\sum_{i,j} \left\| \mathbf{r}_{\text{loop}}^{i,j} \right\|^2_{\boldsymbol{\Lambda}_{\text{loop}}^{i,j}} 
+ \sum_{i} \left\| \mathbf{r}_{\text{aerial}}^{i} \right\|^2_{\boldsymbol{\Lambda}_{\text{aerial}}^{i}} \notag \\
&+ \sum_{i} \left\| \mathbf{r}_{\text{imu}}^{i,i+1} \right\|^2_{\boldsymbol{\Lambda}_{\text{imu}}^{i,i+1}} 
+ \sum_{i} \left\| \mathbf{r}_{\text{gnss}}^{i} \right\|^2_{\boldsymbol{\Lambda}_{\text{gnss}}^{i}}\\
&+ \sum_{i} \left\| \mathbf{r}_{\text{odom}}^{i,i+1} \right\|^2_{\boldsymbol{\Lambda}_{\text{odom}}^{i,i+1}},
\end{align}
where each $\boldsymbol{\Lambda}$ denotes the information matrix associated with the corresponding residual term.

Loop closure constraints are introduced by detecting revisited submap pairs via an intersection-over-union (IoU) threshold. Local 3D features, including the Intrinsic Shape Signatures (ISS) keypoints and Binary Shape Context (BSC) descriptors, are used to estimate the relative transformation $\left( \hat{\mathbf{R}}_{i,j}, \hat{\mathbf{t}}_{i,j} \right)$. The resulting residual is:
\begin{align}
\mathbf{r}_{\text{loop}}^{i,j} =
\begin{bmatrix}
\log \left( \hat{\mathbf{R}}_{i,j}^{-1} \left( \mathbf{R}_i^{-1} \mathbf{R}_j \right) \right)\\
\mathbf{R}_i^{-1} \left( \mathbf{t}_j - \mathbf{t}_i \right) - \hat{\mathbf{t}}_{i,j}
\end{bmatrix}.
\end{align}

Aerial-ground constraints are derived by registering MLS submaps to ALS point clouds using Iterative Closest Point (ICP). The alignment residual is:
\begin{align}
\mathbf{r}_{\text{aerial}}^{i} =
\begin{bmatrix}
\log \left( \hat{\mathbf{R}}_{\text{aerial}}^{-1} \mathbf{R}_i \right) \\
\mathbf{t}_i - \hat{\mathbf{t}}_{\text{aerial}}
\end{bmatrix}.
\end{align}

Odometry constraints model the relative motion between consecutive LiDAR frames:
\begin{align}
\mathbf{r}_{\text{odom}}^{i,i+1} =
\begin{bmatrix}
\log \left( \hat{\mathbf{R}}_{i,i+1}^{-1} \left( \mathbf{R}_i^{-1} \mathbf{R}_{i+1} \right) \right) \\
\mathbf{R}_i^{-1} \left( \mathbf{t}_{i+1} - \mathbf{t}_i \right) - \hat{\mathbf{t}}_{i,i+1}.
\end{bmatrix}.
\end{align}

IMU pre-integration factors are introduced to capture inertial constraints between LiDAR scans. The residual is defined as:
\begin{align}
\mathbf{r}_{\text{imu}}^{i,i+1} &=
\begin{bmatrix}
\delta \boldsymbol{\alpha}_{i+1}^{i} \\
\delta \boldsymbol{\beta}_{i+1}^{i} \\
\delta \boldsymbol{\gamma}_{i+1}^{i} \\
\delta \mathbf{b}^a \\
\delta \mathbf{b}^g
\end{bmatrix},\\
&=
\left[
\begin{aligned}
&\mathbf{R}_i^{-1} \left( \mathbf{t}_{i+1} - \mathbf{t}_i - \mathbf{v}_i \Delta t + \tfrac{1}{2} \mathbf{g} \Delta t^2 \right) - \hat{\boldsymbol{\alpha}}_{i+1}^{i} \\
&\mathbf{R}_i^{-1} \left( \mathbf{v}_j - \mathbf{v}_i + \mathbf{g} \Delta t \right) - \hat{\boldsymbol{\beta}}_{i+1}^{i} \\
&\log \left( \left( \hat{\boldsymbol{\gamma}}_{i+1}^{i} \right)^{-1} \mathbf{R}_i^{-1} \mathbf{R}_j \right) \\
&\mathbf{b}_{i+1}^a - \mathbf{b}_i^a \\
&\mathbf{b}_{i+1}^g - \mathbf{b}_i^g
\end{aligned}
\right].
\end{align}
Here, $\hat{\boldsymbol{\alpha}}_{i+1}^{i}$, $\hat{\boldsymbol{\beta}}_{i+1}^{i}$, and $\hat{\boldsymbol{\gamma}}_{i+1}^{i}$ are the pre-integrated position, velocity, and rotation increments, respectively. $\mathbf{g}$ denotes the gravity vector, and $\Delta t$ is the integration time interval.

GNSS constraints incorporate RTK-based absolute position measurements with lever-arm compensation:
\begin{align}
\mathbf{r}_{\text{gnss}}^i = \mathbf{t}_i + \mathbf{R}_i \cdot \mathbf{t}_{\text{ant}} - \hat{\mathbf{t}}_i^{\text{GNSS}},
\end{align}
where $\mathbf{t}_{\text{ant}}$ represents the lever arm from the platform center to the GNSS antenna.

The overall nonlinear least squares problem is solved via the Levenberg--Marquardt algorithm~\citep{R101}, with computational efficiency enhanced by sparse matrix factorization using SuiteSparse~\citep{R102}. Figure~\ref{fig_pgo} illustrates the pose graph optimization framework. The MLS point clouds are segmented into multiple submaps, each constrained by four types of factors: the aerial-ground association factor, the GNSS factor, the IMU pre-integration factor, and the LiDAR odometry factor. Loop closure factors are introduced to establish connections between submaps with revisited areas.

\section{Ground truth evaluation}
\label{sec_5}
To assess the accuracy and effectiveness of the generated ground-truth trajectories, both qualitative and quantitative evaluations are conducted. Three representative scenes are selected for evaluation: the medium dataset in UrbanNav~\citep{R54}, the Bay Bridge dataset from UrbanLoco~\citep{R60}, and the Loop 1 dataset in Wuhan.

\subsection{Qualitative evaluation}
\begin{figure*}
\centering
\includegraphics[width=0.9\textwidth, keepaspectratio]{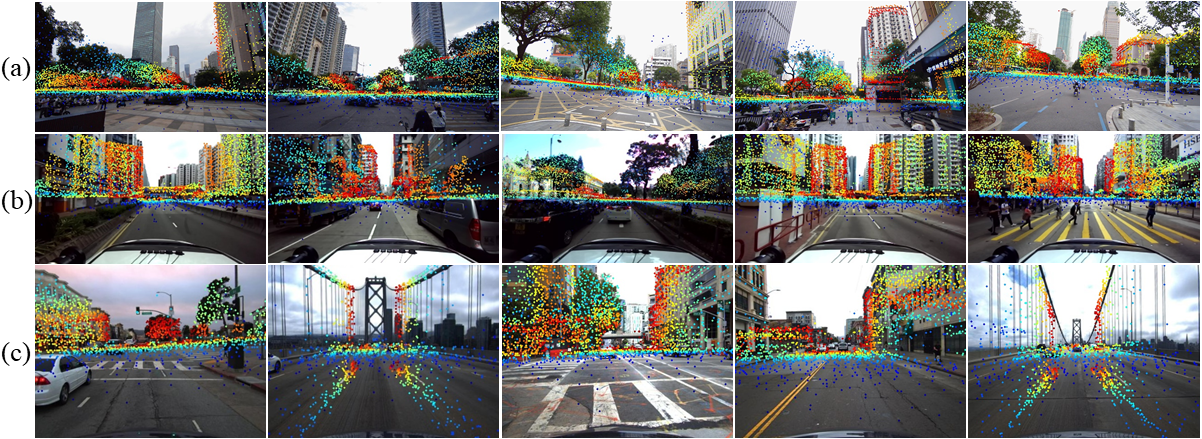}
\caption {Projection of ALS point clouds to images. (a), (b) and (c) are from Wuhan, Hong Kong and California datasets, respectively. Point clouds are colorized by depth, with colors ranging from blue (near) to red (far), through green and yellow.}
\label{fig_l2i}
\end{figure*}

\begin{figure}[]
\centering
\includegraphics[width=\linewidth, keepaspectratio]{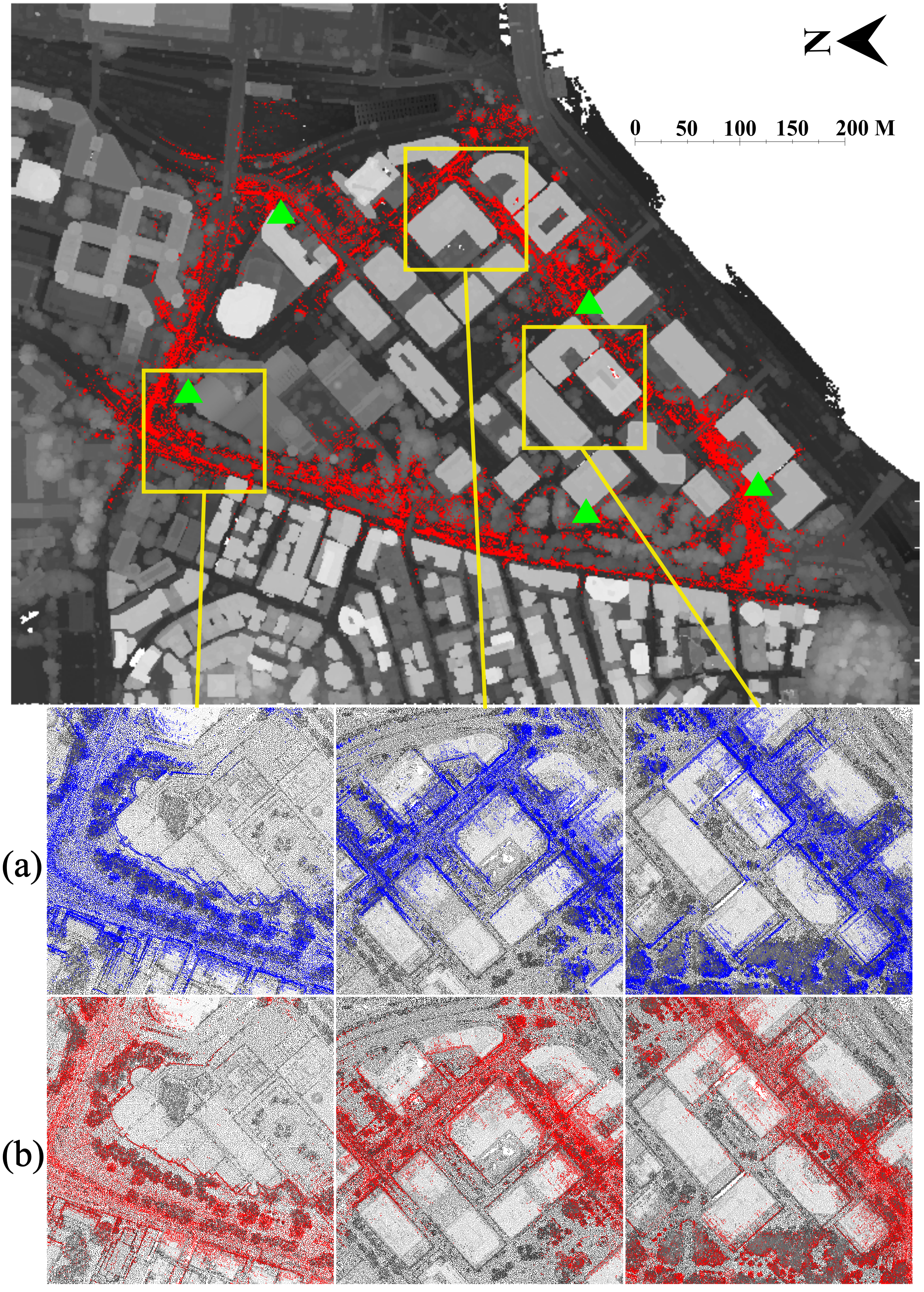}
\caption{Point cloud refinement of the Hong Kong Medium dataset. (a) and (b) illustrate the MLS point clouds before and after optimization. ALS points are rendered in grayscale to represent relative elevation, while green triangles indicate the check points.}
\label{fig_hk}
\end{figure}

\begin{figure*}
\centering
\includegraphics[width=0.9\textwidth, keepaspectratio]{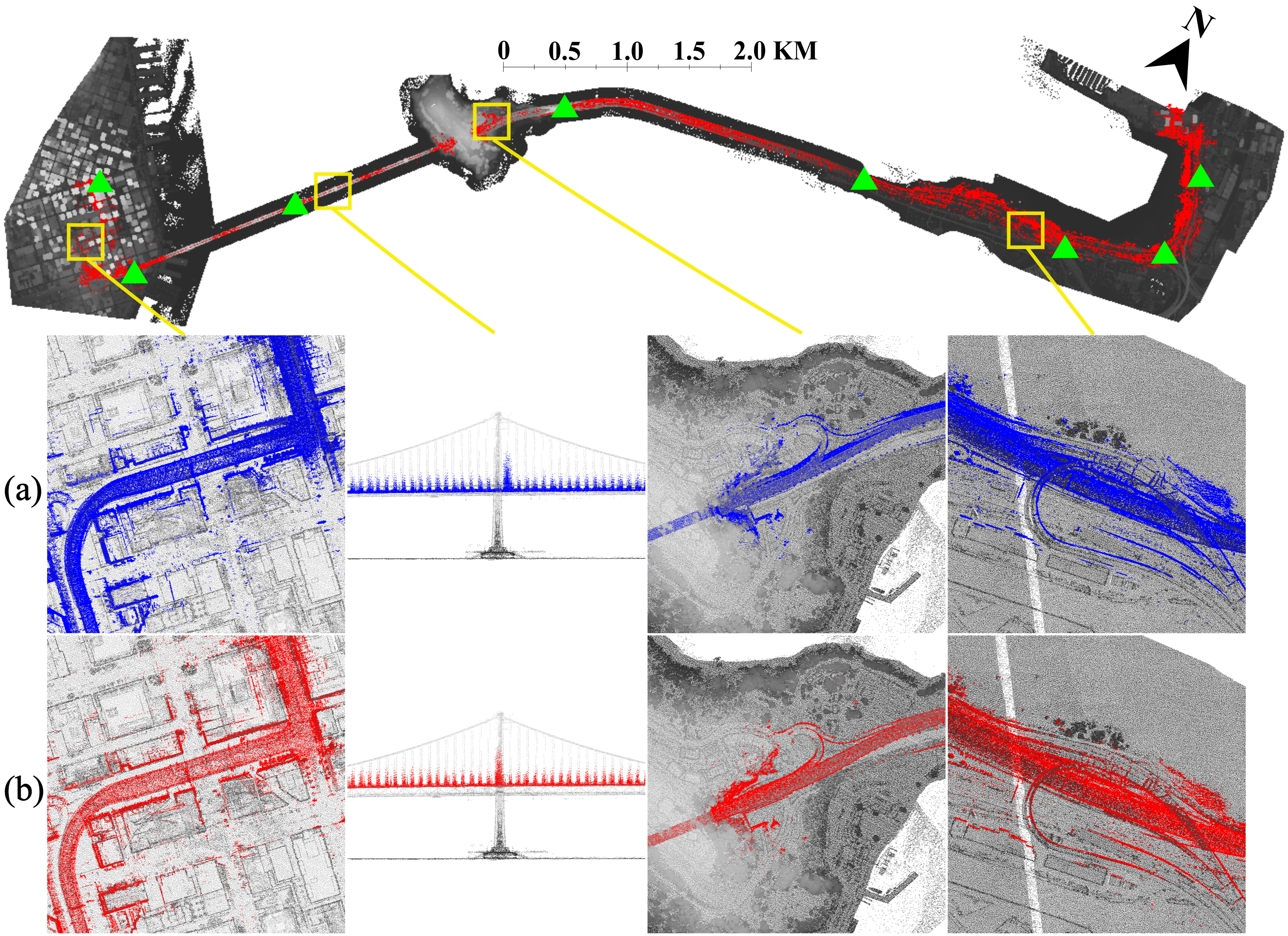}
\caption {Point cloud refinement of California Bay Bridge dataset. (a) and (b) illustrate the MLS point clouds before and after optimization. ALS points are rendered in gray to represent relative height, while green triangles indicate the check points.}
\label{fig_ca}
\end{figure*}

\begin{figure}
\centering
\includegraphics[width=\linewidth, keepaspectratio]{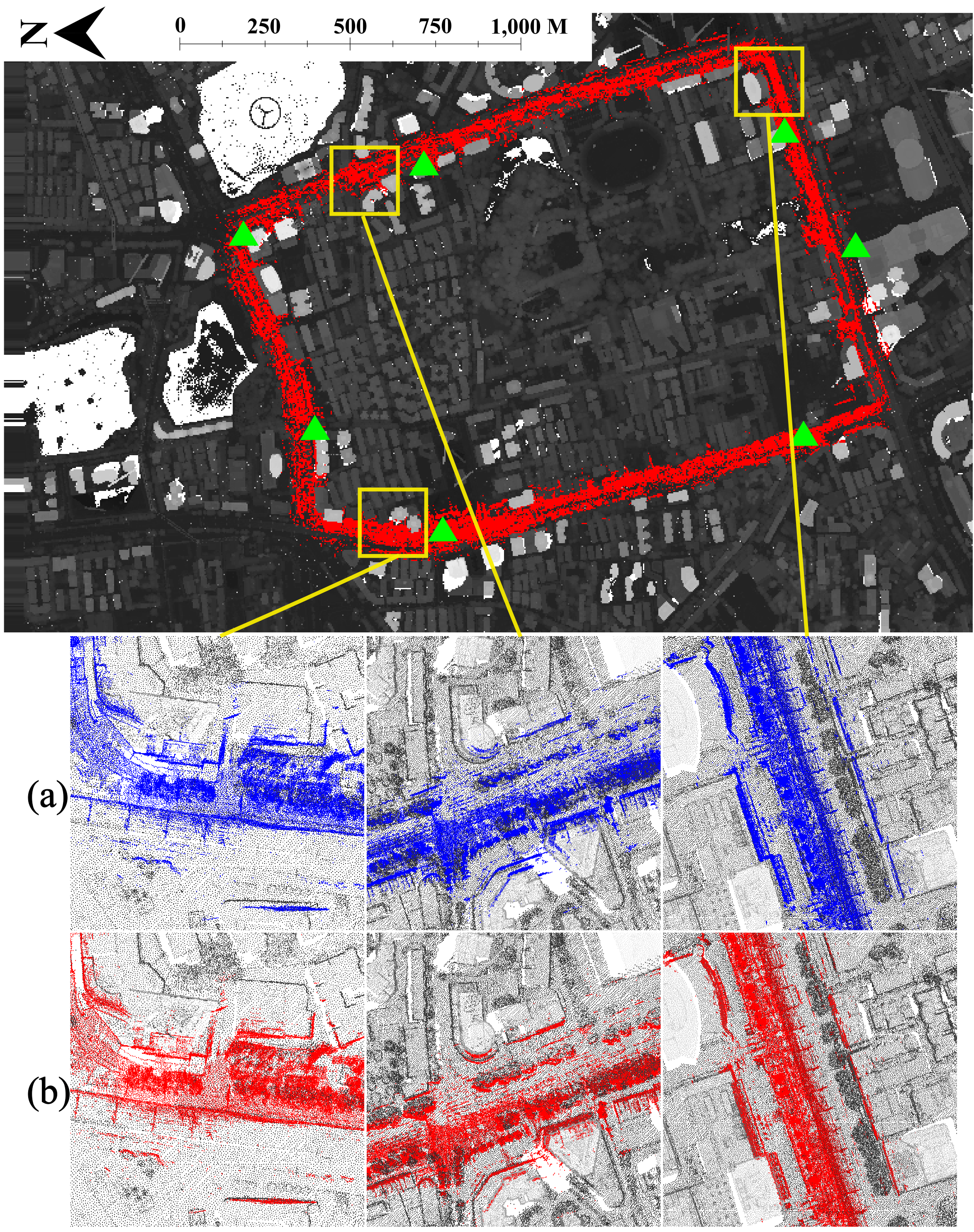}
\caption{Point cloud refinement of the Wuhan Loop 1 dataset. (a) and (b) illustrate the MLS point clouds before and after optimization. ALS points are rendered in grayscale to represent relative elevation, while green triangles indicate the check points.}
\label{fig_wh}
\end{figure}

To qualitatively evaluate the optimized trajectories, we project the ALS point clouds onto the vehicle cameras. This approach offers an intuitive means of assessing the ground-truth trajectory. Representative scenarios of the three datasets are displayed in Figure ~\ref{fig_l2i}. Visualizations show a well-aligned projection, where both geometric structures and depth cues are clearly consistent. Besides, the improved MLS point cloud quality demonstrates the effectiveness of the trajectory generation. As shown in Figures~\ref{fig_hk}, \ref{fig_ca}, and \ref{fig_wh}, the MLS point clouds are better aligned with the ALS point clouds, where blue and red points represent the point clouds before and after optimization, respectively.

\subsection{Quantitative evaluation}
As shown in Figures ~\ref{fig_hk}, \ref{fig_ca}, and \ref{fig_wh}, several checkpoints were manually selected and aligned between the MLS and ALS point clouds to evaluate the accuracy of the trajectory. These significant feature points are primarily located at the corners of pedestrian overpasses and buildings. According to \ref{tab_check_point_errors}, in the dense urban datasets of Hong Kong Medium and Wuhan Loop 1, the average checkpoint errors are 11 cm and 9 cm, respectively. In contrast, the California Bay Bridge dataset yields a higher average error of 0.16 m due to its long span. Details of check point errors are presented in Table ~\ref{tab_check_point_errors}.

\begin{table}[!t]
\centering
\caption{Check point errors with generated ground truth trajectories.}
\label{tab_check_point_errors}
{\small
\begin{tabular}{lccc}
\toprule
\multirow{2}{*}{Sequence} & \multicolumn{3}{c}{Check point error (m)} \\ \cline{2-4}
                          & Avg. & Min. & Max. \\ \midrule
California Bay Bridge     & 0.16 & 0.11 & 0.23 \\
Hong Kong Medium          & 0.11 & 0.07 & 0.15 \\
Wuhan Loop 1              & 0.09 & 0.03 & 0.12 \\
\bottomrule
\end{tabular}
}
\end{table}

\section{Benchmarks on SOTA algorithms} \label{sec:Benchmarks}
\begin{figure*}[!t]
\centering
\includegraphics[width=0.9\textwidth, keepaspectratio]{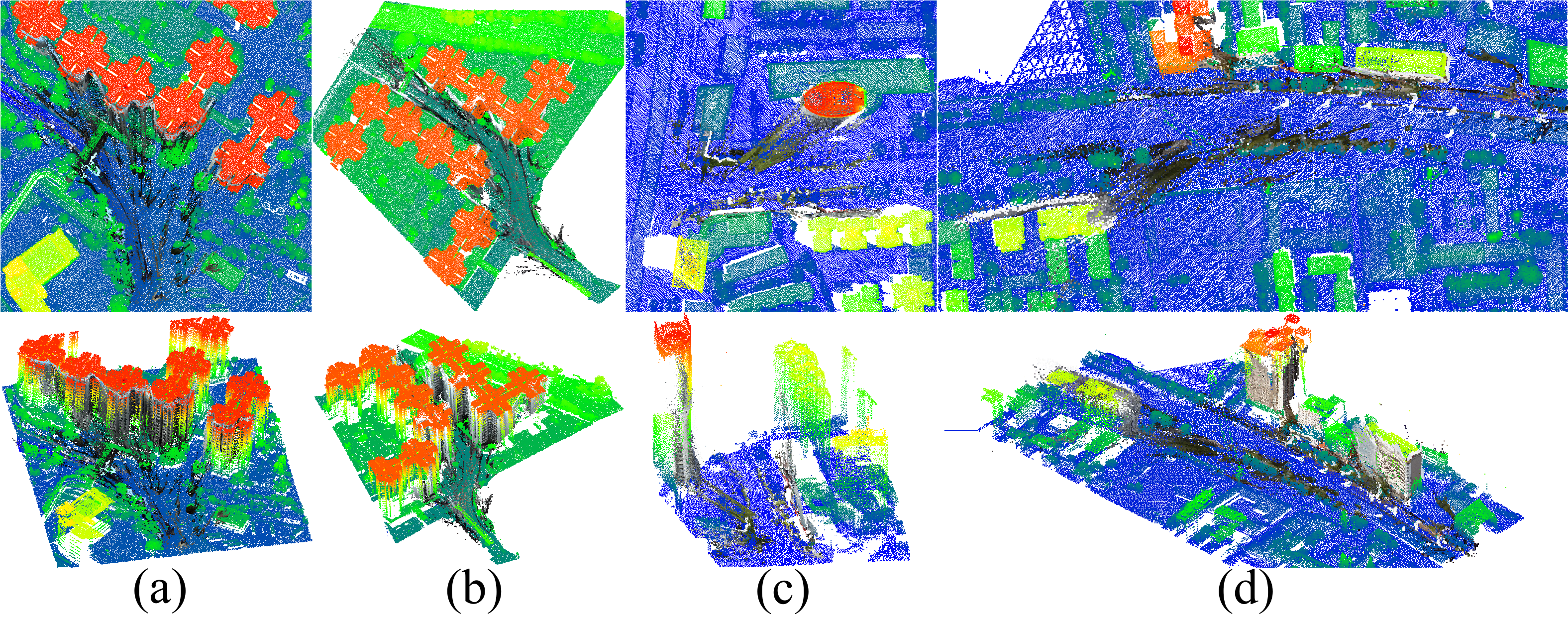}
\caption {I2P fine localization results. Dark points are from dense matching of images, and ALS point clouds are colorized by height.}
\label{fig_i2p}
\end{figure*}

\label{sec_6}
\subsection{I2P global localization}

Three learning-based baseline methods were selected to evaluate the performance of image-to-point cloud (I2P) global localization on our dataset, including:

\textbf{AE-Spherical}~\citep{zhaoAttentionEnhancedCrossmodalLocalization2023}, is the first to achieve end-to-end cross-modality visual localization. It uses point clouds generated by multi-beam vehicle-borne LiDAR point clouds as the map representation and panoramic images as input, aligning multi-modality data through a feature representation learning architecture. In the image branch, AE-Spherical employs a spherical CNN-based backbone for feature extraction, followed by NetVLAD~\citep{arandjelovicNetVLADCNNArchitecture2018} aggregation to obtain global descriptors. In the point cloud branch, it directly employs PointNetVLAD~\citep{uyPointNetVLADDeepPoint2018} to generate global features. Furthermore, the channel attention~\citep{huSqueezeandExcitationNetworks2018}  is incorporated into both branches to enhance feature representation and improve cross-modality alignment performance. We modified the feature backbone by replacing the spherical CNN with ResNet and maintaining other modules unchanged. We retrained their model\footnote{\url{https://github.com/Zhaozhpe/AE-CrossModal}} on our datasets.

\textbf{LIPLoc}~\citep{shubodhLIPLocLiDARImage2024}, which first projects the point cloud into a depth map and then aligns the RGB image and the depth map in a high-dimensional feature space using a contrastive learning paradigm, thereby achieving cross-modal coarse visual localization. Specifically, the point cloud frames are converted into range images based on the scan pattern of the LiDAR sensor. The input images are resized to $224 \times 224$, and then a visual transformer-based feature encoder along with a linear projection layer is employed to generate global features for both the images and the range images. We retrained their model\footnote{\url{https://github.com/Shubodh/lidar-image-pretrain-VPR}} using their default parameters on our datasets.

\textbf{SaliencyI2PLoc}~\citep{liSaliencyI2PLocSaliencyguidedImage2025}, a novel contrastive learning-based architecture that integrates saliency maps into the feature aggregation process and preserves feature relationship consistency across multiple manifold spaces. To reduce the need for extensive data mining in the preprocessing stage, a contrastive learning framework is employed to effectively establish cross-modality feature mappings. A context saliency-guided local feature aggregation module is proposed to exploit stationary scene information, thereby generating more representative global features. Additionally, to further improve cross-modal feature alignment during contrastive learning, the model incorporates the consistency of relative sample relationships across different manifold spaces. We retrained their model\footnote{\url{https://github.com/whu-lyh/SaliencyI2PLoc}} on our datasets.

Following the SaliencyI2PLoc~\citep{liSaliencyI2PLocSaliencyguidedImage2025}, the recall value is utilized to evaluate the I2P global localization quantitative performance. For the California dataset, we select the 0411 and 4701 sequence to assess the performance, while the part of Deep sequence of Hong Kong dataset and Loop-1 sequence of Wuhan dataset are chosen to be the evaluation dataset, respectively. We establish almost 14k and 15k image-point cloud pairs from Hong Kong and Wuhan dataset to train the model. The evaluation dataset is not contained in the training dataset. The AE-Spherical baseline requires the construction of positive and negative sample pairs, we collected the query image every 3 $m$ and set the radius distance to be 20 $m$ and 40 $m$ to classify the positives and negatives. The detail of the dataset splitting is listed in Table~\ref{tab_dataset_i2p_global_loc}.

\begin{table}[!t]
\renewcommand{\arraystretch}{1.3}	
\caption {The dataset details.}
\label{tab_dataset_i2p_global_loc}
\resizebox{0.48\textwidth}{!}{
\begin{tabular}{llllllllll}
\toprule
&\multicolumn{3}{c}{CA}  & \multicolumn{3}{c}{HK}  & \multicolumn{3}{c}{HK} \\ \cline{2-10}
& Query & Database & Total & Query & Database & Total & Query & Database & Total
\\\midrule 
Train           & -       & -      & -      & -         & -         & 14100     & -         & -         & 15551     \\ 
Evaluation      & -       & -      & -      & 206       & 2280      & 2486      & 664       & 3180      & 3844     \\ 
\bottomrule
\end{tabular}%
}
\end{table}

The quantitative results on the California, Hong Kong, and Wuhan datasets are presented in Table~\ref{recall_ca} and Table~\ref{recall_hk_wuhan}. LIPLoc~\citep{shubodhLIPLocLiDARImage2024} employs projected images as proxies to bridge the modality gap between images and point clouds, thereby achieving the best performance across all evaluated datasets. In contrast, both SaliencyI2PLoc~\citep{liSaliencyI2PLocSaliencyguidedImage2025} and AE-Spherical~\citep{zhaoAttentionEnhancedCrossmodalLocalization2023} directly operate on raw point clouds. Due to differences between airborne and vehicle-mounted point clouds—such as variations in point density, noise characteristics, and spatial distribution—these two methods demonstrate limited effectiveness in cross-modality localization. Although SaliencyI2PLoc slightly outperforms AE-Spherical, particularly on the Hong Kong and Wuhan sequences, LIPLoc consistently achieves superior results across all metrics. Given the broader coverage and higher acquisition efficiency of airborne LiDAR compared to vehicle-mounted systems, vision-based localization using airborne point clouds offers greater potential for real-world applications. Nonetheless, the current performance of both SaliencyI2PLoc and AE-Spherical remains far from practical usability, highlighting substantial room for future improvement.


\begin{table}[!t]
\centering
\caption{I2P global localization recall values on the California dataset.}
\label{recall_ca}
\resizebox{0.48\textwidth}{!}{
\begin{tabular}{llllllll}
\toprule
\multirow{2}{*}{Method} & \multicolumn{3}{c}{0411 sequence} & \multicolumn{3}{c}{4706 sequence} \\ \cline{2-7}
                        & R1$(\%)\uparrow$ & R5$(\%)\uparrow$ & R20$(\%)\uparrow$ & R1$(\%)\uparrow$ & R5$(\%)\uparrow$ & R20$(\%)\uparrow$ \\
\midrule
AE-Spherical~\citep{zhaoAttentionEnhancedCrossmodalLocalization2023}    & 8.76  & 13.87 & 16.79 & 24.36 & 30.77 & 35.90 \\
LIPLoc~\citep{shubodhLIPLocLiDARImage2024}                              & 13.14 & 30.66 & 47.45 & 16.67 & 34.62 & 53.85 \\
SaliencyI2PLoc~\citep{liSaliencyI2PLocSaliencyguidedImage2025}          & 8.03  & 11.68 & 16.06 & 19.23 & 23.08 & 24.36 \\
\bottomrule
\end{tabular}
}
\end{table}

\begin{table}[!t]
\centering
\caption{I2P global localization recall values on the Hong Kong dataset and Wuhan dataset.}
\label{recall_hk_wuhan}
\resizebox{\linewidth}{!}{
\begin{tabular}{llllllll}
\toprule
\multirow{2}{*}{Method} & \multicolumn{3}{c}{Deep sequence} & \multicolumn{3}{c}{Loop-1 sequence} \\ \cline{2-7}
                        & R1$(\%)\uparrow$ & R5$(\%)\uparrow$ & R20$(\%)\uparrow$ & R1$(\%)\uparrow$ & R5$(\%)\uparrow$ & R20$(\%)\uparrow$ \\
\midrule
AE-Spherical~\citep{zhaoAttentionEnhancedCrossmodalLocalization2023}   & 10.68 & 14.56 & 17.96 & 3.61  & 5.87  & 8.43  \\
LIPLoc~\citep{shubodhLIPLocLiDARImage2024}                             & 19.90 & 35.92 & 60.19 & 16.50 & 35.44 & 54.85 \\
SaliencyI2PLoc~\citep{liSaliencyI2PLocSaliencyguidedImage2025}         & 20.87 & 22.82 & 24.27 & 6.48  & 7.68  & 10.09 \\
\bottomrule
\end{tabular}
}
\end{table}

\subsection{I2P fine localization}

To address fine-grained localization between images and point clouds, we evaluated three representative learning-based I2P registration methods: DeepI2P~\citep{li2021deepi2p}, CorrI2P~\citep{ren2022corri2p}, and CoFiI2P~\citep{kang2024cofii2p} with their official implementations, adapted and retrained on our dataset. A brief overview of each method is provided below.

\textbf{DeepI2P}~\citep{li2021deepi2p}: reformulates image-to-point cloud pose estimation as a two-stage classification and optimization problem. Instead of learning explicit  descriptors for point cloud and image, it uses a two-branch classification network with attention-based image–point cloud fusion to predict whether each 3D point lies inside the camera frustum, thus reducing reliance on handcrafted keypoints and feature matching. A inverse camera projection solver then recovers the relative camera-to-LiDAR transformation from the classification results. This design bypasses direct point-to-pixel correspondence estimation, minimizes storage, and remains robust under challenging viewpoint and modality gaps. We retrained the official model\footnote{\url{https://github.com/lijx10/DeepI2P}} on our dataset.

\textbf{CorrI2P}~\citep{ren2022corri2p}: treats the image-to-point cloud registration as a dense correspondence estimation task. It employs a dual-branch architecture for modality-specific feature extraction and a multi-scale fusion mechanism to progressively integrate contextual cues. A confidence-guided correspondence regression module improves match reliability, while a transformation estimation layer jointly refines rotation and translation parameters. This design enhances robustness under large viewpoint changes and modality gaps. We retrained the official model\footnote{\url{https://github.com/rsy6318/CorrI2P}} on our dataset.

\textbf{CoFiI2P}~\citep{kang2024cofii2p}: adopts a coarse-to-fine strategy, combining global pose initialization with local refinement. The coarse stage estimates an approximate transformation from global context features, while the fine stage iteratively refines the pose using local geometry and semantic cues. A confidence-guided correspondence filtering mechanism suppresses noisy matches, balancing efficiency and accuracy in complex scenes. We retrained the official model\footnote{\url{https://github.com/WHU-USI3DV/CoFiI2P}} on our dataset.

Despite careful adaptation, none of these methods succeeded in producing reliable correspondences or accurate relative poses in our data. We attribute this primarily to the lack of structured objects in our scenes (e.g., poles, traffic signs, other salient landmarks) that are crucial for cross-modal matching. In contrast to image–LiDAR pairs acquired simultaneously by the same platform, our aerial-based point clouds and vehicle-based images suffer from substantial viewpoint and acquisition-time discrepancies, further exacerbating the registration challenge.Specifically, DeepI2P, which depends on dense pixel-to-point similarity learning, is highly sensitive to the absence of distinctive geometric structures, making feature associations in sparse, texture-less regions ambiguous. CorrI2P, which relies on local correspondence matching guided by global priors, fails when repetitive or weak visual cues dominate, leading to unstable matches. CoFiI2P, which depends on accurate coarse alignment, accumulates large errors in its initial stage under severe viewpoint discrepancies, leaving the fine stage ineffective.

After the failure of learning-based approaches, we perform Structure-from-Motion (SfM) on images sampled along the trajectory at intervals of 100 meters and recover the scale using VINS-Mono. Subsequently, we conduct ICP registration between the dense matching image point clouds and the ALS point clouds. Figure~\ref{fig_i2p} demonstrates the I2P fine localization results. The registration accuracy is assessed using Relative Rotation Error (RRE) and Relative Translation Error (RTE). The Euler angle vector $\mathbf{r}$ is derived from the relative rotation matrix $\mathbf{R}_{gt}^{-1} \mathbf{R}_e$, where $\mathbf{R}_{gt}$ and $\mathbf{t}_{gt}$ denote the ground-truth rotation and translation and $\mathbf{R}_e$ and $\mathbf{t}_e$ denote the estimated rotation and translation.

\begin{align}
\text{RRE} = \sum_{i=1}^{3} \left| \mathbf{r}_i \right|, 
\quad
\text{RTE} = \left\| \mathbf{t}_{gt} - \mathbf{t}_{e} \right\|_2.
\end{align}

\begin{table}[!t]
\centering
\caption{I2P fine localization accuracy on different regions.}
\label{tab_i2p_fine}
{\small
\resizebox{\linewidth}{!}{
\begin{tabular}{lcc}
\toprule
Region        & $\mathrm{RRE}(^{\circ})$ & $\mathrm{RTE}(\mathrm{m})$ \\ \midrule
Hong Kong     & $4.06 \pm 3.29$          & $2.87 \pm 2.08$            \\
Wuhan         & $7.80 \pm 2.21$          & $2.64 \pm 1.25$            \\
California    & $14.54 \pm 11.82$        & $4.59 \pm 4.14$            \\
\bottomrule
\end{tabular}
}}
\end{table}

The quantitative results for different regions are summarized in Table~\ref{tab_i2p_fine}. In Hong Kong dataset, the method achieves the best accuracy, with an RRE of $4.06^\circ$ and an RTE of $2.87$ meters. Wuhan shows slightly higher rotational error ($7.80^\circ$) but a similar translational error ($2.64$ meters). California exhibits the largest errors among the three regions, with an RRE of $14.54^\circ $ and an RTE of $4.59 $ meters. In particular, the California dataset contains cross-sea bridges and dense urban areas (e.g. Chinatown), where repetitive structures and occlusions further complicate image matching.

\section{Challenge issues and future directions}
\label{sec_7}
Our experimental results underscore the limitations of current state-of-the-art methods when applied to large-scale, real-world, and cross-platform scenarios. In the following, we highlight key open problems in the field and discuss potential research directions to advance the robustness, scalability, and practicality of cross-modal localization systems.

\subsection{Modality discrepancy and cross-domain generalization}

As evidenced in Section \ref{sec:Benchmarks}, existing I2P models such as AE-Spherical and SaliencyI2PLoc suffer from degraded performance when applied to cross-modal and cross-view scenarios, especially in the California dataset. This is largely due to the large modality gap between perspective ground-view images and top-down ALS point clouds, compounded by viewpoint and appearance variations. 

Many current models still depend heavily on modality-specific features—such as texture patterns in images or height variations in point clouds—which limits their ability to generalize across modalities. Addressing this issue calls for approaches that can extract features robust to both modality and domain changes. Promising directions include: (1) domain-adaptive learning to reduce the domain shift between different modalities and environments; (2) learning shared latent spaces that preserve structural and semantic correspondences; and (3) disentangling geometric information from appearance variations to enhance generalizability across modalities and environments.

\subsection{Structural inconsistency and scene complexity}

Table~\ref{tab_i2p_fine} shows higher localization errors in the California areas, which are dominated by either open-span infrastructure or extremely dense urban blocks. Both scenarios pose significant challenges for cross-modal alignment: the former provides sparse and repetitive geometry with few stable constraints, while the latter presents highly cluttered scenes with severe occlusions, making both dense matching and ICP registration difficult. In contrast, structured urban areas such as Hong Kong and Wuhan, offer denser and more distinctive geometry, leading to more stable matching. This highlights the sensitivity of both learning-based and registration-based methods to geometric completeness and façade continuity. Future research could explore: (1) incorporating semantic segmentation to identify stable structural elements; (2) applying façade completion techniques to recover missing geometry; and (3) leveraging CAD-aligned priors to introduce precise geometric constraints, thereby improving registration robustness in sparse or ambiguous environments.

\subsection{Temporal misalignment and appearance change}

Current I2P pipelines face significant challenges when aligning datasets collected across different times. In our case, ALS data acquired by governmental agencies and ground imagery from mobile mapping systems are often collected years apart, leading to temporal discrepancies caused by new construction or demolition, vegetation growth or removal, and seasonal appearance changes. These changes affect both geometry and visual context, making cross-modal correspondence less reliable in both global retrieval and fine registration stages. Future directions include: (1) developing robust feature extractors that identify stable elements across long time spans; (2) adopting change-aware alignment frameworks such as time-conditioned descriptors; and (3) integrating temporal priors (e.g., historical GIS data) to make cross-modal localization more resilient in real-world applications.

\subsection{Computational efficiency and deployment readiness}

Although learning-based methods demonstrate potential, most require high-end GPUs and long inference times, limiting their applicability in real-time or embedded systems. For example, SaliencyI2PLoc achieves higher recall in the Hong Kong sequence but is computationally intensive, making it less suitable for real-time deployment. Future work could focus on: (1) applying model compression techniques, such as pruning and knowledge distillation, to reduce computational cost; (2) adopting hardware-aware training strategies; and (3) developing edge inference frameworks, thereby enabling efficient and reliable on-device deployment of aerial-ground localization systems in robotics, AR/VR, and wearable applications.

\section{Conclusion}
\label{sec_8}

We present a new benchmark for aerial-ground cross-modal localization, integrating mobile mapping imagery with ALS point clouds across three major cities. To support accurate evaluation, we propose an indirect ground-truth generation pipeline based on MLS–ALS alignment and pose graph optimization. Experimental results reveal that existing I2P methods still face significant challenges under large-scale, cross-view, and cross-modal conditions, while ALS-based localization offers promising accuracy and stability. Our benchmark provides a standardized platform for evaluating global and fine-grained localization tasks, aiming to advance research in robust, scalable visual localization under real-world urban scenarios.

\printcredits

\section{Declaration of interests}
The authors declare that they have no known competing financial interests or personal relationships that could have appeared to influence the work reported in this paper.

\bibliographystyle{cas-model2-names}

\bibliography{ref}

\end{sloppypar}
\end{document}